# Bioacoustical Periodic Pulse Train Signal Detection and Classification using Spectrogram Intensity Binarization and Energy Projection

**Marian Popescu** CP478@CORNELL.EDU
**Peter J. Dugan** PJD78@CORNELL.EDU
**Mohammad Pourhomayoun** MP749@CORNELL.EDU
Bioacoustics Research Program (BRP), Cornell University, Ithaca, NY, USA, 14850

**Denise Risch** DENISE.RISCH@NOAA.GOV
Northeast Fisheries Science Center, Woods Hole, MA, USA, 02543

**Harold W. Lewis III** HLEWIS@BINGHAMTON.EDU
Department of Systems Science and Industrial Engineering, Binghamton University, NY, USA, 13850

**Christopher W. Clark** CWC2@CORNELL.EDU
Bioacoustics Research Program (BRP), Cornell University, Ithaca, NY, USA, 14850

## Abstract

The following work outlines an approach for automatic detection and recognition of periodic pulse train signals using a multi-stage process based on spectrogram edge detection, energy projection and classification. The method has been implemented to automatically detect and recognize pulse train songs of minke whales. While the long term goal of this work is to properly identify and detect minke songs from large multi-year datasets, this effort was developed using sounds off the coast of Massachusetts, in the Stellwagen Bank National Marine Sanctuary. The detection methodology is presented and evaluated on 232 continuous hours of acoustic recordings and a qualitative analysis of machine learning classifiers and their performance is described. The trained automatic detection and classification system is applied to 120 continuous hours, comprised of various challenges such as broadband and narrowband noises, low SNR, and other pulse train signatures. This automatic system achieves a TPR of 63% for FPR of 0.6% (or 0.87 FP/h), at a Precision (PPV) of 84% and an F1 score of 71%.

## 1. Introduction

Passive acoustic monitoring allows the exploration of marine mammal acoustic ecology at diverse temporal and spatial scales. While this technique is effective in understanding and characterizing habitats (Clark et al., 1996), it can often generate large acoustical data volumes. Furthermore, the acoustical signal domain presents various challenges such as: non-stationary and non-Gaussian noise, low signal to noise ratio (SNR), self-induced broadband and narrowband sensor noise, abiotic, environmental noise such a rain fall, ice and wind (Martin et al., 2012), and anthropogenic noise caused by vessels (Parks et al., 2009) or seismic airgun exploration activities (Guerra et al., 2011). Therefore, the current research is focused on creating efficient, robust automatic algorithms that can mine, identify, and classify marine mammal sounds across highly variable, large data sets.

Machine learning is an important step in the development of automatic acoustic species detection. Early automatic detection techniques used matched filters, hidden Markov model, and spectrogram cross-correlation (Clark et al. 1987). These methods were later improved through the use of machine learning approaches such as a feed-forward neural network classifier (Mellinger and Clark, 1993; Potter et al., 1994; Deecke et al., 1999; Mellinger, 2004; Mazhar et al., 2007; Pourhomayoun et al., 2013). Other machine learning algorithms, such as classification and regression tree classifiers (CART), have also been implemented in recognizing contact calls made from the North Atlantic Right Whale (Dugan et al., 2010). Improvements over single recognition methods have been shown by using an advanced technique, which combines several recognition methods running in parallel (Dugan et al., 2010; Pourhomayoun et al., 2013).

In this paper we discuss an automated approach, for detecting and classifying periodic, broadband, pulsed signals using machine learning techniques. In particular, we will focus on the detection and classification of minke whale (*Balaenoptera acutorostrata)* songs, and the development of a system that can be applied to other datasets without re-training.

### 1.1 Minke whale (*Balaenoptera acutorostrata)*

The minke whale is a marine mammal species within the suborder of baleen whales and is found throughout the North Atlantic Ocean. Like all whales, minkes use sound to feed, breed, navigate and communicate (Richardson et al., 1995). Recent studies have shown that their perception of sound (Brkic et al., 2004) can be influenced by various environmental conditions such as wind and ice, but also anthropogenic noises (Martin et al., 2012). Therefore, quantifying large-scale biological phenomena such as seasonal occurrence and season distribution is critical for understanding the potential influences of natural and manmade factors on population dynamics. While various minke whale studies have been conducted (Schweder et al., 1997; Oswald et al., 2011), little information is available regarding the North Atlantic minke whale's seasonal distribution and occurrence off the U.S. East Coast. The methodology described here was developed to analyze large data sets collected by Cornell University using Marine Autonomous Recording Units (MARUs) during 2006-2010 (Calupca et al., 2000). The multi-channel data, continuously recorded at 2 kHz, was captured off the coast of Massachusetts, in the Stellwagen Bank National Marine Sanctuary (SBNMS). The algorithm was applied to 895 continuous days in order to analyze the seasonal distribution and occurrence of minke whales (Risch et al., 2000) in the SBNMS.

### 1.2 Signal characteristics and challenges

The minke whale vocalizations are characterized as pulse trains that can last somewhere between 40-60 sec, typically within the 100-1400 Hz frequency band. The pulse trains are comprised of individual pulses lasting 40-60 msec, and can exhibit variable pulse rates ranging from 2.8 pulses/sec to 4.5 pulses/sec (Mellinger et al., 2000). While our proposed methodology can be used for any pulse train series, here we focused on pulse trains contained within the 75-350 Hz frequency band, with variable length Inner Pulse Interval (IPI) described above. Figure 1 depicts the spectrogram of a minke whale pulse train song, as well as additional sources of noise and energy. The challenge is to detect and classify these pulse train signatures as they occur within a continuous stream of acoustic data.

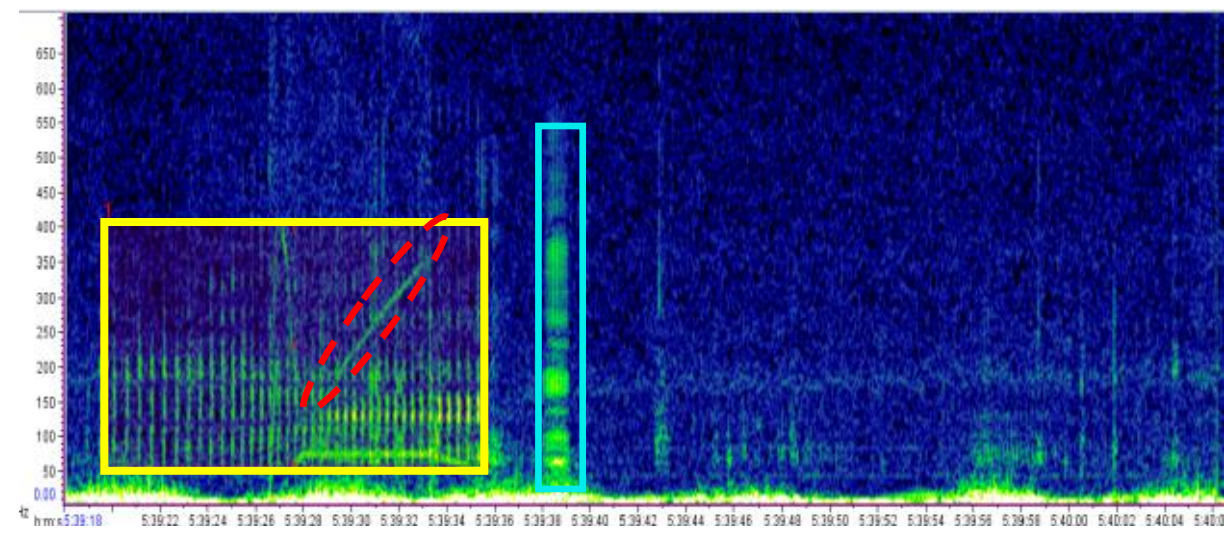

*Figure 1*. The spectrogram of a minke whale vocalization lasting ≈ 17 secs. The yellow box indicates the minke pulse train signature with the variable IPI. The noise generated by hard disk drive (red dotted ellipse) can be seen clearly within the minke pulse train. The spectrogram also reveals energy from an additional species known as Haddock (blue box), constant narrowband noises between 70-200 Hz, other sources of short impulse broadband and low-frequency noises. These noise characteristics change from sensor to sensor and sometimes on a minute by minute basis.

### 1.3 Train and Test Datasets

Since the signal of interest contains such broad variability, a training dataset was created in order to capture the parameter space. The dataset contains 2429 minke pulse trains from each of the 10 sensors. The minke pulse trains were identified, by an expert human biologist, by manually hand browsing randomly chosen subsets of the recordings. Additionally, a total of 2788 noise events that ranged from ambient noise, to shipping vessel noise, sensor hard-drive noise, and other cross species, was added. Overall, the train dataset consists of 112 continuous hours recording and is used in designing the detector and qualitatively analyzing the performance of various classifiers.

Furthermore, in order to analyze the performance of the trained system, a test dataset was created. The test dataset consists of 120 continuous hours, containing 729 total minke vocalizations. The dataset is constructed by using 3 days from Stellwagen Bank National Marine Sanctuary recording and 2 days from other external sensors from the Long Island, New York area. This will allow us to measure how well the methodology can be generalized using the trained model. The test dataset also contains various challenges, including very low SNR vocalizations and as well as additional species know has haddock which also has broadband pulse signals. Figure 2 presents some of the challenges in the test dataset.

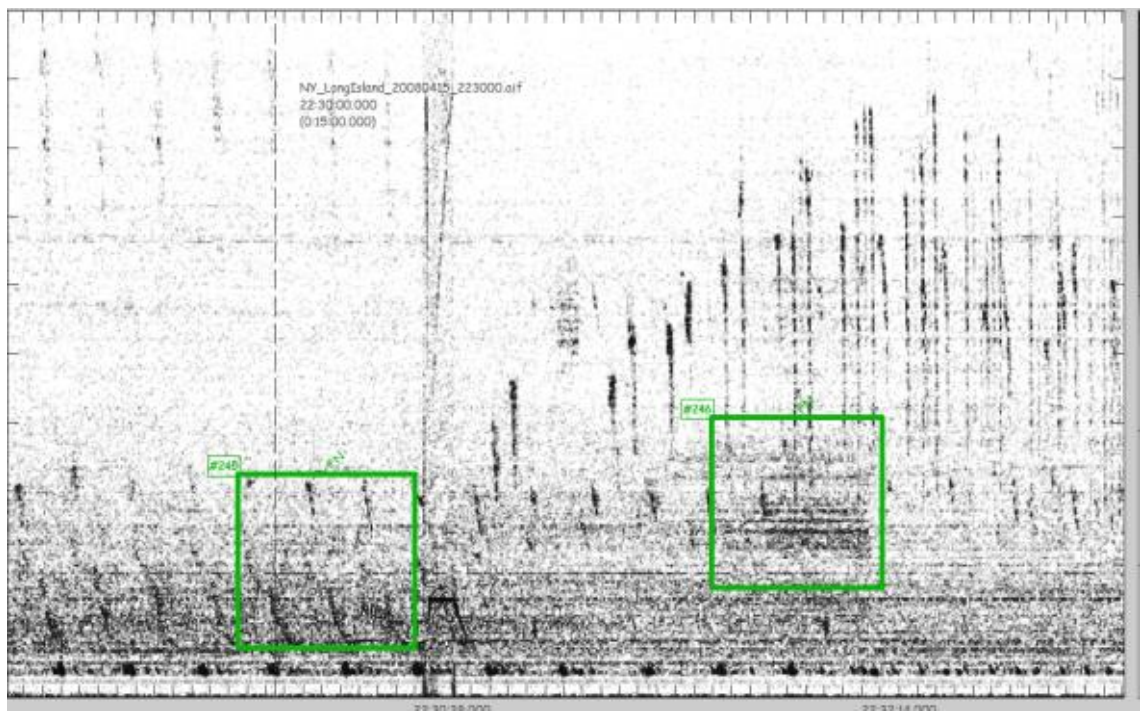

(a)

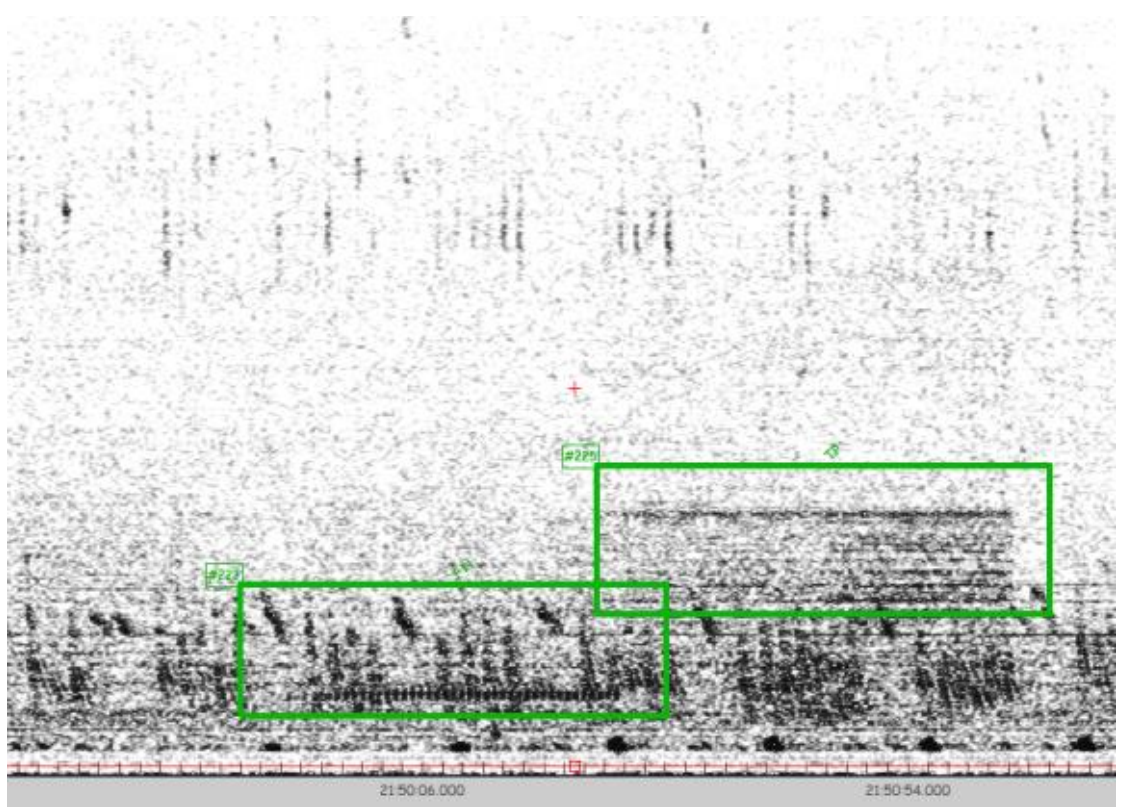

(b)

*Figure 2*. The spectrogram of minke whale vocalizations in the test dataset: (a) low SNR minke vocalization in the left green box, and minke vocalization influenced by other species and broadband pulses in the right box. Other sources noise can be also observed. (b) minke vocalizations superimposed by pulse train signatures created by the haddock species.

## 2. Methods

Previous methods for detecting pulse type vocalizations are based on: (1) cross-correlation with a pre-designed kernel, or (2) auto-correlation of a given signal block (Mellinger and Clark, 1993). However, their performance is highly depended on choice of kernel and threshold. The implementation can also suffer from high computational complexity. The proposed methodology for automatic detecting and classifying of minke pulse trains in a continuous dataset consists of a two-stage approach. In the first stage, we try to identify the pulse train signatures based on a set of rules that match a description of the minke whale signal. In the second stage, we extract a set of features from the detected events, which will be later used to recognize the events using a previously trained classifier.

### 2.1 Stage I – Detecting pulse train signatures

The proposed detection stage consists of several steps. First, since the acoustical data are continuous, a sliding window of duration equal to 30 sec was applied to create the time–domain signal slices s(t). Secondly, since the signals of interest are located within the 75-300 Hz frequency band, s(t) is conditioned using a type II, Chebyshev bandpass FIR filter; with -30 dB attenuation, 40 Hz roll-off, and 0.1 dB of ripple in the passband. The filter is implemented in order to reduce the energy outside the desired frequency bands and to improve the intensity-based spectrogram binarization step. Next, a spectrogram is computed for the filtered s(t) signal using a Blackman window, 8% overlap, 512 point FFT, to yield 20.5 ms time and 3.89 Hz frequency bins. The spectrogram is then cropped to match the frequency band bounds of the bandpass filter. Once the spectrogram is obtained, a binarization based on image intensity is applied in order to denoise the signal and remove the ambient noise, and place the signal in the same basis across all the sensors. First, we convert the spectrogram matrix to a gray-scaled intensity image.

We then compute an intensity mask using:

$$\gamma = 1.75 * \sigma_s + \mu_s \tag{1}$$

where $\mu_s$ is the mean intensity of the image and $\sigma_s$ is the standard deviation of the zero-mean intensity image. The level was derived based on the idea that the signal is not wide-sense stationary, which implies a different mean for each signal slice s(t), and that any acoustical signatures above the mean ambient noise level is captured within the standard deviation. Applying the level masking produces a binarized image, in which all pixels of the gray-scaled image with luminance greater than the level $\gamma$ have a value of 1 (white), and replaces all other pixels with the value 0 (black). Using the $N$x$M$ binarized image matrix, an image energy project function, $P(n)$ is created as:

$$P(n) = \sum_{m=1}^{M} BW(\mathrm{n},\mathrm{m}) \quad \text{for} \quad n = 1, 2, \ldots, N \tag{2}$$

This process will place emphasis on broadband signatures, since pulse spectrogram time slices will contain a large number of vertical pixels (i.e. energy). Next, we find the local maxima of the energy projection function and apply the following set of rules, which have been designed for the minke vocalization pulse train, but can be generalized to any other pulse train signature: (1) local maxima above a threshold; (2) minimum and maximum number of local maxima above the threshold; (3) a range for the local maxima spacing (based on IPI). Any events that meet these criteria are then identified as minke pulse trains and sent to the next stage for feature extraction and classification. Figure 3 illustrates the detection process.

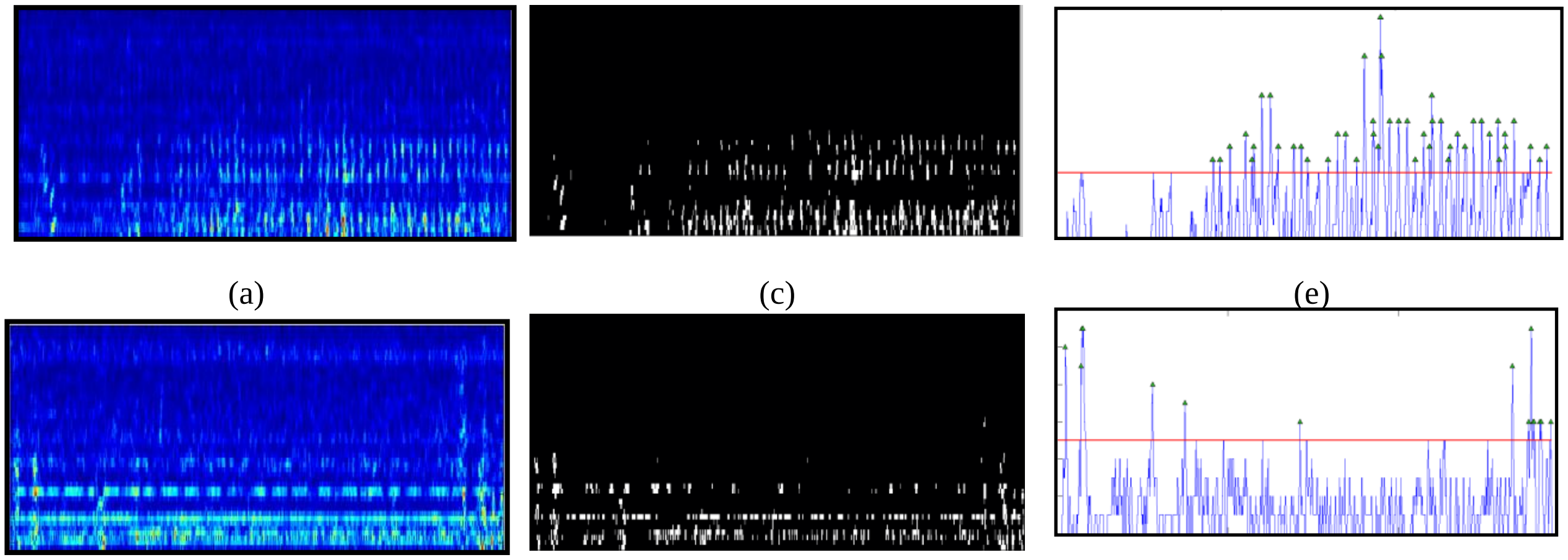

(a) (c) (e)

(b) (d) (f)

*Figure 3*. The detection process for a minke pulse train (top) and noise event (bottom),respectively; (a) and (b), spectrogram after bandpass filtering and cropping, respectively; (c), (d) the intensity-based binarization of the spectrograms, respectively. (e), (f) the energy projection function $P(n)$ with the same applied threshold.

### 2.2 Feature extraction

A set of 18 features is extracted for each detected event. The features are designed and chosen with the intent to distinguish the detected minke pulse trains from the ambient noise events (detector errors). A summary of the selected features is shown in Table 1.

*Table 1*. Features used to train and evaluated the classifiers.

| Feature Number | Feature Name | Description (of Pulse Train) |
|---|---|---|
| F1 | delta time | Duration of pulse train |
| F2-F3 | frequency pair min-max | Frequency bounds |
| F4 | number of clicks | Number of pulses |
| F5 | average bandwidth | Average bandwidth of pulse train |
| F6 | center frequency | Center bandwidth of the pulse train |
| F7 | average sharpness | F4 / F1 |
| F8 | CEC for signal | LEQ of the detected pulses within the pulse train |
| F9 | Mean Leq | Mean LEQ of the detected pulses |
| F10 | DeltaT- mean | The mean of the IPI of detected clicks |
| F11 | DeltaT- mode | The mode of IPI of detected clicks |
| F12 | DeltaT- max | The max IPI of detected clicks |
| F13 | DeltaT- min | The min IPI of detected clicks |
| F14 | SNR | Signal to Noise Ratio of the detected pulse train |
| F15 -18 | SNR: $x^{th}$ percentile | SNR of pulse train using the $5^{th}$ , $10^{th}$ , $20^{th}$ and $25^{th}$ percentile of slice as noise |

### 2.3 Classification

The detection method, discussed above, identifies areas of energy that meet the criterion presented in figure 3; we will refer to these as regions of interest (ROI's). Many of the ROI's which are recognized by the detection stage result from various noise conditions such as vessel noise, or additional marine mammal vocalizations, and thus a classification stage is implemented to increase the overall performance of the system. This stage is designed to reduce the false positive rate of the detector, since in bio-acoustical applications, the analysts have to manually verify the output results. In order to analyze the performance of various classifiers, a feature vector is extracted after applying the detection stage on the train data. Our analysis investigates the performance of the following classifiers: (1) grafted C4 tree with a confidence factor of 0.25 (Webb, 1999), (2) a Random Forest with 10 random trees in the forest and 5 features used in random selection (Breiman, 2001), (3) a Bayesian network via a Simple Estimator with alpha equal to 0.5 and K2 search algorithm (Cooper and Herskovits, 1992), a ripple-down rule learner with 3 fold used for pruning and 2 minimum weights of the instances in a rule (Gaines and Compton ,1992) and a functional tree that did not use binary split and used 15 instances for node splitting (Gama, 2004; . The methods are evaluated using at a 66%, 33% split on the training data. The performance of the classifiers is shown in Figure 4. It can be seen that the random forest classifier has the best area under the curve (AUC).

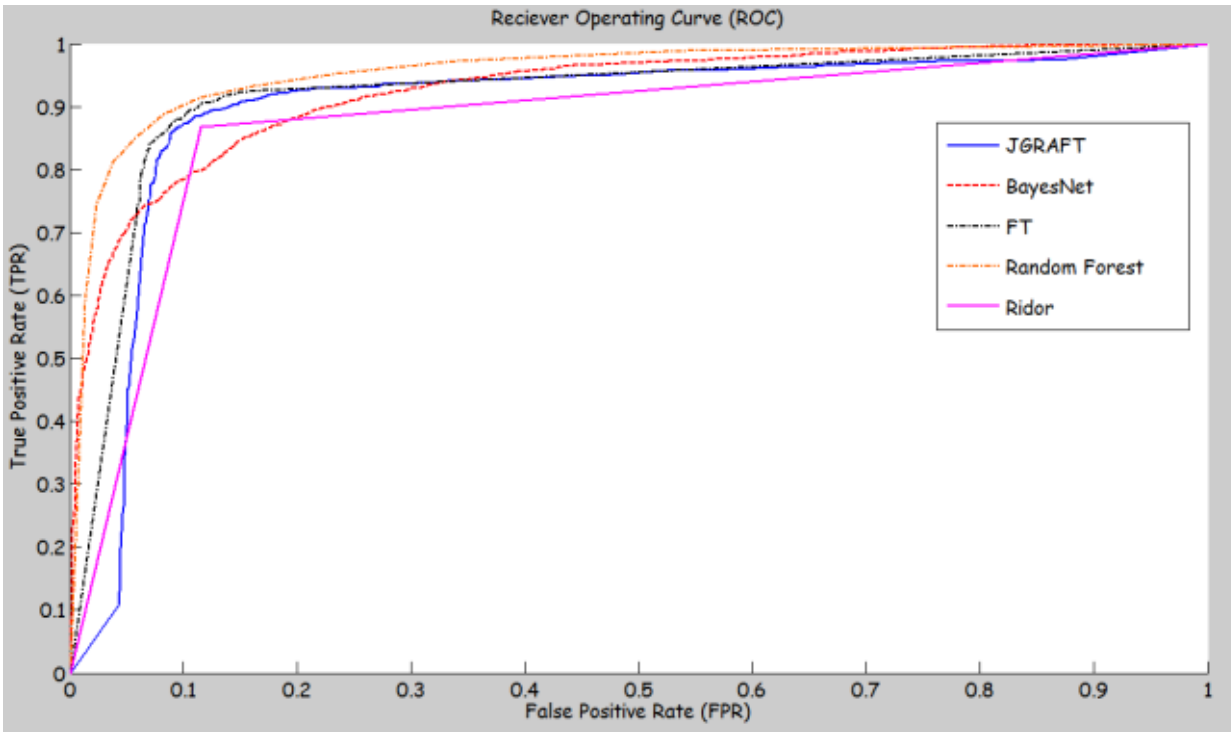

*Figure 4*. The performance of various classifiers across the training dataset, using a 66% split for training and a 33% split for test. The figure and area under the curve indicate that the random forest classifier is the best option for our given feature space.

## 3. Results and Conclusion

The proposed technique was applied on a test dataset using an energy projection function with threshold equal to 6. A total number of 28820 signal slices, of which 3158 were minke vocalizations, were analyzed by the detector. The detection stage produces a True Positive Rate (TPR) of 79%, a False Positive Rate (FPR) of 11% or 15.48 False Positives per hour (FP/h), at a Precision (PPV) of 40% and an F1 score of 53%. In order to reduce the number of false positives generated by the detector, a random forest classifier is applied on the testing dataset. The performance of the proposed classifier on the testing dataset is shown below in Table 2.

*Table 2*. The performance of the trained classifier on the challenge test data without further training.

| TPR | FPR | Precision | F1 | AUC | Class |
|---|---|---|---|---|---|
| 94% | 36% | 84% | 0.89 | 85% | Non-Minke |
| 79% | 6% | 84% | 0.72 | 85% | Minke |

It can be seen that the performance of the classifier diminished when applied to the new testing dataset. This was due to the low SNR conditions, and other interfering broadband signatures that were being detected. If increased performance in true positive is required, the signal should either be further de-noised, additional features should be added to the training data, or the training vector size should be increased to include detection events from the test data. When the detector and trained classifier system is applied to the test data, it produced a TPR of 63% for FPR of 0.6% (0.87 FP/h), at a PPV of 84% and an F1 score of 72%. It should be noted that while the TRP went from 79% to 63%, the FPR went from 11% to 0.6%.

In this paper we have shown the design and implementation of an automatic detection and classification system, used to mine and identify minke whale pulse trains within a continuous stream of acoustic data. The results show that the proposed method can achieve high performance even in the presence of high noise conditions.